\documentclass[lettersize,journal]{IEEEtran}
\usepackage{amsmath,amsfonts}
\usepackage{algorithmic}
\usepackage{algorithm}
\usepackage{array}
\usepackage[caption=false,font=normalsize,labelfont=sf,textfont=sf]{subfig}
\usepackage{textcomp}
\usepackage{stfloats}
\usepackage{url}
\usepackage{paralist}
\usepackage{graphicx}
\usepackage{cite}
\usepackage[dvipsnames]{xcolor}
\usepackage{multirow}
\usepackage{tikz}
\usetikzlibrary{shadows.blur}

\newcommand{\ie}{\textit{i}.\textit{e}.,~}
\newcommand{\eg}{\textit{e}.\textit{g}.,~}
\newcommand{\etal}{\textit{et~al}.~}
\newcommand{\wrt}{\textit{w}.\textit{r}.\textit{t}.~}
\newcommand{\cf}{\textit{cf}.~}

\newcommand\copyrighttext{%
\small \copyright 2024 IEEE.  Personal use of this material is permitted.  Permission from IEEE must be obtained for all other uses, in any current or future media, including reprinting/republishing this material for advertising or promotional purposes, creating new collective works, for resale or redistribution to servers or lists, or reuse of any copyrighted component of this work in other works.
DOI 10.1109/TIV.2024.3428415
}
\newcommand\copyrightnotice{%
	\begin{tikzpicture}[remember picture,overlay]
	\node[anchor=north,yshift=-10pt] at (current page.north) {\fbox{\parbox{\dimexpr\textwidth-\fboxsep-\fboxrule\relax}{\copyrighttext}}};
	\end{tikzpicture}%
}

\begin{document}

\title{Deep Learning Safety Concerns in Automated Driving Perception}

\author{Stephanie Abrecht, Alexander Hirsch, Shervin Raafatnia, Matthias Woehrle
	\thanks{Stephanie Abrecht, Alexander Hirsch and Shervin Raafatnia are with Cross-Domain Computing Solutions, Robert Bosch GmbH, Stuttgart, Germany,  (email: \{stephanie.abrecht, alexander.hirsch, shervin.raafatnia\}@de.bosch.com)}
	\thanks{Matthias Woehrle is with Corporate Research, Robert Bosch GmbH, Renningen, Germany. (email: matthias.woehrle@de.bosch.com)}
	\thanks{Alphabetical order of authors.}
}


\maketitle
\copyrightnotice

\begin{abstract}
Recent advances in the field of deep learning and impressive performance of deep neural networks (DNNs) for perception have resulted in an increased demand for their use in automated driving (AD) systems. The safety of such systems is of utmost importance and thus requires to consider the unique properties of DNNs.
In order to achieve safety of AD systems with DNN-based perception components in a systematic and comprehensive approach, so-called safety concerns have been introduced as a suitable structuring element. 
On the one hand, the concept of safety concerns is -- by design -- well aligned to existing standards relevant for safety of AD systems such as ISO 21448 (SOTIF). 
On the other hand, it has already inspired several academic publications and upcoming standards on AI safety such as ISO PAS 8800.
While the concept of safety concerns has been previously introduced, this paper extends and refines it, leveraging feedback from various domain and safety experts in the field. 
In particular, this paper introduces an additional categorization for a better understanding as well as enabling cross-functional teams to jointly address the concerns.
\end{abstract}

\begin{IEEEkeywords}
Deep Learning, Automated Driving, Safe Perception, Safety-critical systems
\end{IEEEkeywords}

\maketitle

\section{Introduction}
\label{sec:intro}

Deep learning approaches have shown remarkable performance across perception, prediction, and planning tasks.
As such, deep neural networks (DNNs) are widely used in AD systems, especially in perception. 
In such safety-critical automated systems, a detailed understanding of the impact of DNNs on overall system safety is of utmost importance.
The focus is on the safety of the intended functionality (SOTIF), in scope of ISO~21448~\cite{sotif}, of an otherwise fault-free system.
To this end, this paper discusses the concept of safety concerns of DNNs, introduced in~\cite{willers2020safety}, as a suitable structuring element for a systematic and comprehensive analysis.

\begin{figure}
	\centering
	\includegraphics[width=\columnwidth]{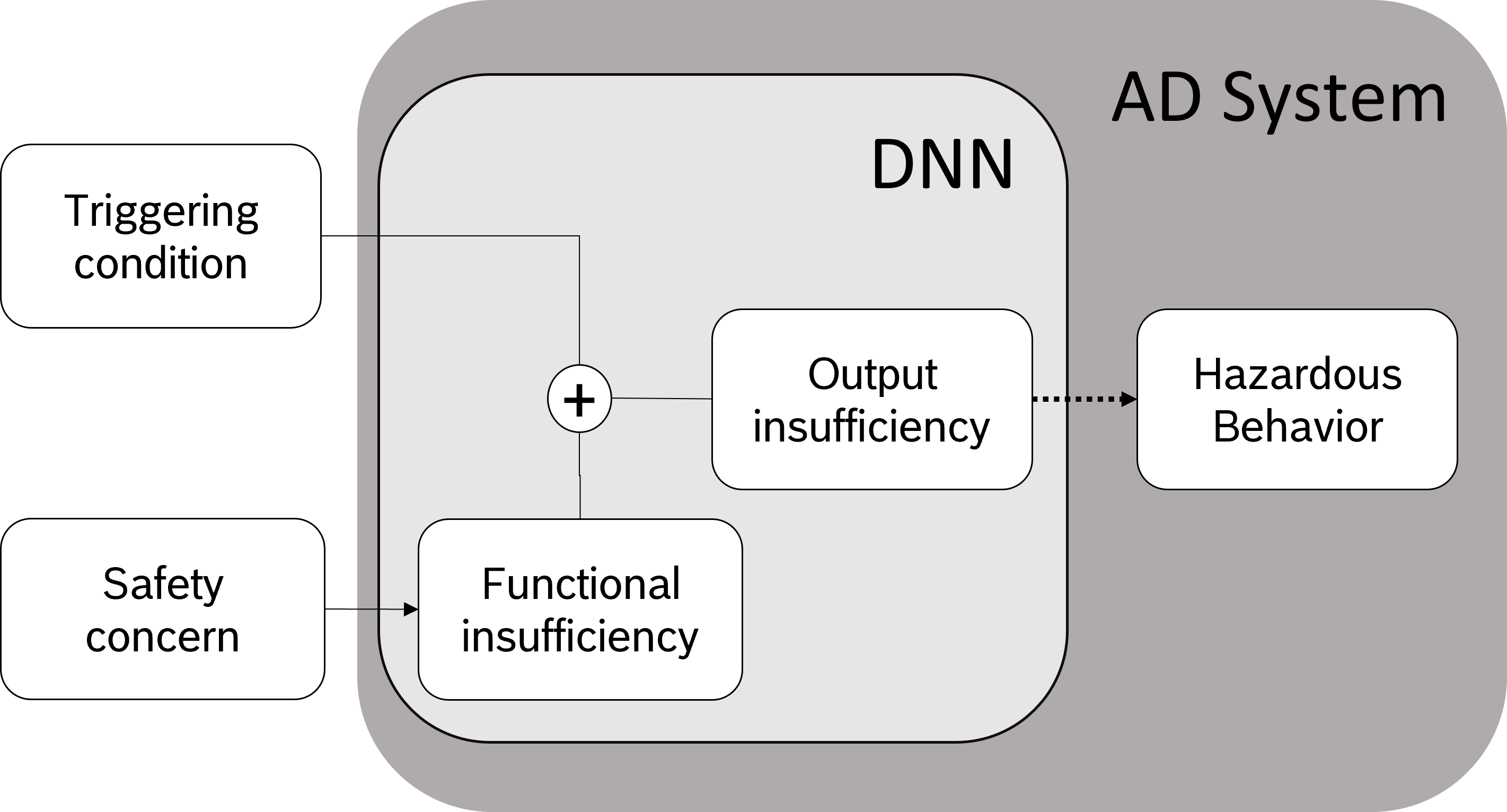}
	\label{fig:intro}
	\caption{The terminology used in this paper, which is aligned with ISO 21448 (SOTIF)~\cite{sotif}: A safety concern can lead to a functional insufficiency within a DNN. Once the functional insufficiency is triggered by a triggering condition, it results in an output insufficiency of the DNN. Output insufficiencies may lead to hazardous behavior of the system.}
\end{figure}

Safety concerns are well aligned to the SOTIF cause and effect model~\cite[Fig. 3]{sotif}.
In the SOTIF cause and effect model, a triggering condition can activate a functional insufficiency of a system element, which may lead to an output insufficiency of this element and subsequently may contribute to hazardous behavior on the vehicle level.
Figure~\ref{fig:intro} shows how this model from ISO~21448 relates to the concept of safety concerns of a DNN-based system element for perception.
Safety concerns are defined as the source of a functional insufficiency of a DNN. 
Such a functional insufficiency may -- once triggered -- result in an output insufficiency of the element, which in turn may lead to hazardous behavior of the AD system.

As an example, consider the task of stopping at an intersection with a stop sign: If a traffic sign detection DNN systematically misses the stop sign (output insufficiency) due to the triggering condition of an adversarial patch in the scene~\cite{metzen2021meta}, the vehicle may not stop, which would be a hazardous behavior.
As we can see, it is vital to understand the sources for output insufficiencies of DNNs.
In the example, we can see that brittleness of DNNs is the source of a functional insufficiency such that an adversarial input (triggering condition) results in a misprediction.

The main motivation of this work is to structure the problem space (``What are the safety concerns?'') to guide future work on the solution space, \ie mitigations (``How can safety concerns be addressed?'').
There are a few reasons for this choice.
First, as we will see in the following, safety concerns can be derived based on the application -- here AD systems -- and the corresponding task of DNNs.
This allows us to introduce an application-specific level of abstraction such that approaches and artefacts relating to safety concerns can be leveraged across projects.
In contrast, mitigations and their evaluation typically require a concrete system and a particular task description and cannot be transferred to other use cases without adaptations.

Second, addressing output insufficiencies directly -- if possible at all -- will not be sufficient, because depending on the underlying safety concern, very different triggering conditions may lead to the same output insufficiency and thus require differing mitigations. 
Therefore, it is vital to first focus on safety concerns resulting in functional insufficiencies that lead to an output insufficiency when triggered.
Moreover, there may be many possible mitigations for an individual safety concern. In the adversarial examples case, there are several available mitigations including adversarial training and sensor fusion.
At the same time, mitigations may help to address several underlying issues, \eg sensor fusion may help with brittleness of individual DNNs \wrt adversarial examples or temporal instability of predictions, yet it also supports uncertainty quantification.

We will focus on safety concerns for perception tasks without feedback loop and non-recurrent DNNs trained in a supervised fashion, including semi- and self-supervised variants.
As we outline in this work, these safety concerns for AD systems can be organized into four categories relating to (\textit{i})~the open world the automated vehicle operates in (operational design domain), (\textit{ii})~data and data set preparation, (\textit{iii})~DNN characteristics, and (\textit{iv})~the analysis and evaluation of the DNNs within their operational design domain.

This work provides the following contributions:

\begin{compactenum}
	\item We present a comprehensive and refined list of safety concerns that has evolved from previous work~\cite{willers2020safety} by discussions among safety experts in the field of AD and Safe AI.
	\item This includes a categorization of safety concerns based on the source of the safety concerns, which may originate in the domain, the DNNs and corresponding data, as well as analysis and evaluation.
\end{compactenum}

In contrast to our previous work~\cite{willers2020safety} on safety concerns, we provide the following novel contributions:

\begin{compactenum}
	\item We refactor and complement the original nine safety concerns into fourteen refined safety concerns and detail on this refactoring.
	\item We structure safety concerns into four categories depending on their sources. This also helps addressing them by relevant teams in an organization.
\end{compactenum}

This work is structured as follows.
We first introduce the background of this work in Sec.~\ref{sec:background}.
We then define safety concerns and present a categorization of the safety concerns in Sec.~\ref{sec:concerncategorization}.
We describe all safety concerns within their corresponding category: We start with the \textit{open-world context} in~Sec.~\ref{sec:open}. We continue to \textit{data and data set preparation} concerns in ~Sec.~\ref{sec:data}. Then we show concerns related to \textit{DNN characteristics} in Sec.~\ref{sec:DNN} and finally present \textit{analysis and evaluation} concerns in Sec.~\ref{sec:eval}. 
After presenting related work in Sec.~\ref{sec:related_work}, we conclude the paper.

\section{Background}
\label{sec:background}

While many of the points discussed in this paper could apply to different systems and use cases, it is important to clarify the scope of this work: 
We focus on DNN-based perception for AD systems.
For the sake of simplicity of the presentation, we assume that the system is fixed in the sense that it would not undergo major adaptations, such as adding a new sensor modality or changing the sensor fusion concept, which would change the task or relevance of its components. 
Similarly, the intended usage of the system is also assumed to be fixed and conformed to. For example, driving on mountain roads with an automated vehicle developed for highways is out of scope of this paper.
Moreover, most examples are related to vision-based perception tasks such as pedestrian detection or drivable space characterization.
For a driving task, \eg navigating in an urban area, the environment needs to be perceived in detail, \eg identifying lanes and objects including their class and location.
Therefore, our focus is on DNNs which yield dense predictions, such as segmentation, object detection, optical flow.
This means that tasks such as image classification with a single, global prediction for a datum are out-of-scope of this paper.
This results in a function from a high-dimensional input space to a high-dimensional output space.
As an example, for pixel-wise classification of images with $n$ pixels, $k$ values per pixel, and a segmentation task formulation with $c$ classes, the number of possible functions $\mathbb{R}^{n^{k}} \rightarrow \mathbb{R}^{n^{c}}$ is typically vast - too vast to be comprehended or exhaustively analyzed.

\subsection{Operational Design Domain}
\label{subsec:dist}

A system is designed to operate in the world under specific conditions. This is usually referred to as the operational design domain (ODD) of the system. More precise definitions differ across fields.
In the automotive industry, the commonly accepted definition is provided by the Society of Automotive Engineers (SAE). According to the SAE J3016 (2021) standard, ODD is defined as following for driving automation systems: ``Operating conditions under which a given driving automation system, or feature thereof, is specifically designed to function, including, but not limited to, environmental, geographical, and time-of-day restrictions, and/or the requisite presence or absence of certain traffic or roadway characteristics.''~\cite{sae2021taxonomy} As such, it can be considered as a relatively high-level semantic description of the domain in which an AD system is supposed to function.

For the purposes of this paper, we need to consider the \textit{ODD distribution} $\mathcal{O}_{\mathcal{W}}$, where $\mathcal{W}$ is a set of properties of the world.
Additionally, we need to consider a concrete task with a defined set of labels $\mathcal{Y}$ and input domain $\mathcal{X}$. $\mathcal{Y}$ could be a fixed set of classes of traffic participants, or elements in a world model, and  $\mathcal{X}$  the domain of concrete sensor inputs such as pixels or point clouds.
The \textit{sensor data distribution} of a sensor $k$, $\mathcal{S}^k_{(\mathcal{X}, \mathcal{Y}|\mathcal{W})}$ is conditional on the ODD, and therefore, on the open world. 
While sensor data distributions, \eg across modalities, are typically different they often share the same ODD distribution.
Orthogonal to the kind of distribution - ODD or sensor data - is the concept of whether we consider the population distribution or a sample, \ie the sample distribution.

Up to now we talked about the target population distribution, \ie the actual distribution in the world in the target domain.
However, in practice we only see samples drawn from the population, \ie the sample distribution.
During development, we just have access to the sample distribution in the form of training, validation, and test sets, which we call development data in the following.
Therefore, estimation of model properties, such as its generalization capability, can only be an approximation. 
This approximation is subject to uncertainties, and even worse, may have systematic differences due to sampling issues.
Moreover, sensor measurements are the proxy to elements of interest in the ODD. For example, the cameras provide us with a sensor data distribution of pixels while the distribution of different objects, scenarios, etc., is what is considered in engineering an AD system.   

From our discussion, we can see that there are different distributions of relevance for a safety analysis. This might be the sensor data distribution on sensor feature level, \eg pixels or points clouds, which directly feeds into a deep learning task.
Such a sensor data distribution is conceptually different from a semantically described ODD distribution defined by humans, \eg based on weather, road features and traffic participants.

In deep learning, or more generally machine learning, a basic assumption is that the population distribution is fixed~\cite{von2011statistical}.
However, there are several reasons why this distribution may change.
On the one hand, there can be changes on the sensor data distribution, \eg the sensor itself changes due to aging effects on the sensor. 
On the other hand, since there is a dependency on properties of the world, a changing ODD distribution also impacts the sensor data distribution and results in a distributional shift as further described below.
Note that we separate below two different sources of distributional shift.
One source mainly stems from the evolution of the world over time and is therefore related to the open-world context, \cf~Sec. \ref{subsec:shift}, while the other is with respect to data and its domain and therefore a data and data set specific concern, \cf~Sec. \ref{sub:mismatch}.
Let us consider the introduction of electrical scooters and corresponding drivers:
The introduction of this kind of traffic participant was not foreseen and is not available in many detection datasets and would thus generally be classified as a pedestrian. 

\subsection{Risk in Safety and Machine Learning}
\label{subsec:risk}

Safety and machine learning both feature the concept of risk minimization.
In machine learning, empirical risk minimization (ERM) is used where the underlying distribution is approximated by empirical, independent and identically distributed (i.i.d.), samples~\cite{mohri2018foundations}.
In typical machine learning applications, each datum is equally weighted via standard loss functions, such as cross-entropy loss or mean squared error, and thus provides the same contribution to risk.

This is in contrast to safety-related applications where we know that some cases, and hence data, have higher severity and therefore relevance than others.
A typical characterization of risk is a product of \textit{exposure} and \textit{severity}.
In analogy to ERM, we can see that exposure (frequency) is captured by the underlying distribution and severity needs to be specifically introduced in the loss function, \eg by additional weights.
Such weights can be determined by an analysis of safety relevance, \eg considering a severity for pedestrian detection is described by Lyssenko~\etal~\cite{Lyssenko2022}.
Nevertheless, as common in machine learning, weighting can also be indirectly performed via data samples, \ie creating an ``adjusted exposure'' that considers severity by under- and oversampling data.
In the following, we use the term risk from the safety perspective.

As described above, for ERM we assume i.i.d. sampling, such that these sample distributions are a good approximation of the population.
For such a high-dimensional distribution achieving an (approximate) i.i.d. sample is very difficult.
In particular, we need to consider the tails of the distribution, \cf~\cite{yang2022survey}, especially when events in the long tail have a non-negligible probability.
Outside the ERM realm in machine learning and apart from practical feasibility issues, i.i.d. sampling is not always suitable~\cite{hutchinson2022evaluation}, \eg when evaluating the influence of specific aspects on model performance.

\section{Safety Concerns Categorization}
\label{sec:concerncategorization}

In this section, we first provide a definition of safety concerns.
Subsequently, we detail on the categorization of safety concerns.

\subsection{Definition of Safety Concerns}
\label{subsec:sc_intro}

As discussed in the introduction, in this paper, a \textit{safety concern} is a source of a functional insufficiency in the DNN. Once it is triggered, this functional insufficiency results in an output insufficiency in the corresponding part of the system, as defined in ISO~21448~\cite{sotif}. 
Depending on the actual situation and the error propagation path in the system, an output insufficiency may result in hazardous behavior at system level, as discussed on the SOTIF cause and effect model in Sec.~\ref{sec:intro}. 

As a concrete example, a DNN-based video perception model may fail to generalize to previously unseen data that is within the scope of the system it is embedded in.
In this case, the \textit{triggering condition} could be a previously unseen road sign that mandates stopping.
The \textit{safety concern} here is the DNN not having been trained with similar signs.
The \textit{output insufficiency} is the DNN-based video perception model not recognizing the sign correctly.
This in turn could lead to hazardous behavior which in this case means that the system does not realize it should stop. 
However, if the novel sign is misclassified as another sign which also mandates stopping, the safety concern does not lead to hazardous behavior.

Here, we see that safety concerns increase the likelihood of output insufficiencies at the model level, thus also increasing the likelihood of hazardous behavior of the system: a DNN may predict correctly on a concrete datum never seen before, but continuously providing unseen data to the DNN increases the likelihood of mispredictions.
In this example, one possible mitigation could be monitoring for unseen data during operation. 
Mitigations are used to decrease the risk of hazardous behavior of the AD system. 
The concept of safety concerns as structuring elements helps to demonstrate absence of unreasonable residual risk for the safety of the intended function~\cite{sotif} of DNN-based systems.
Note that in the following, we will discuss in several safety concerns that there may be residual risks due to unknown influences, \eg due to the open-world context.
However, this is not a particular idiosyncrasy of DNN-based systems, but for any AD system and therefore described in standards such as ISO~21448.
ISO~21448 relies on the concept of known or unknown scenarios which can cause hazardous or non-hazardous behavior of an ADS as a structuring element.
Concerning unknown scenarios, \cite[clause 11]{sotif} stipulates that it shall be validated that the residual risks from these are at an acceptable level.
Furthermore, \cite[clause 13]{sotif} mandates field monitoring processes during the operating phase of the ADS in order to monitor the correctness of the estimation and identify new risks resulting from context evolution. 

\subsection{Categorization of Safety Concerns}
\label{subsec:categ}

\begin{figure*}[th]
	\centering
	\includegraphics[width=1.5\columnwidth]{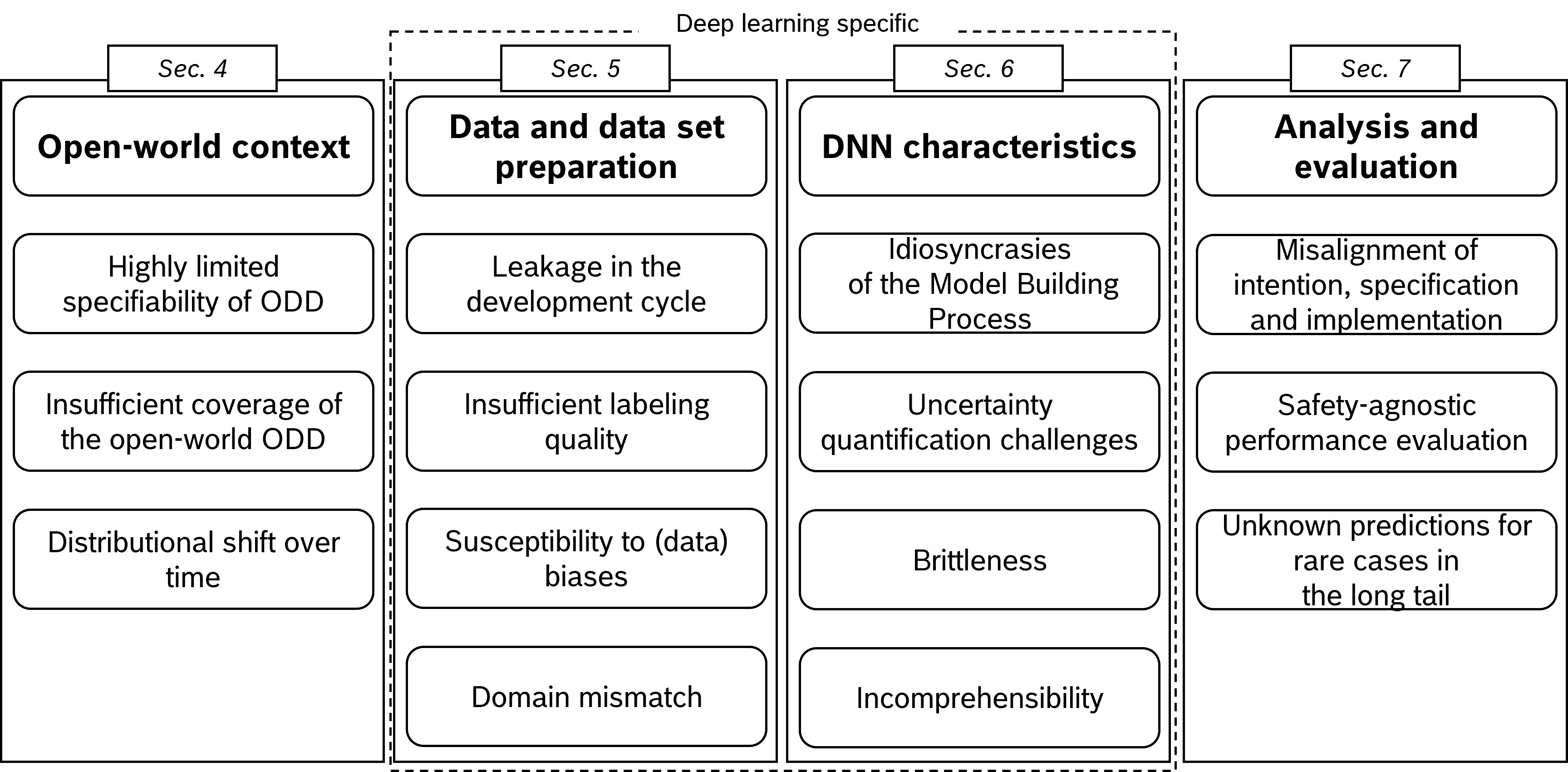}
	\label{fig:structure}
	\caption{Safety concerns categorization overview.}
\end{figure*}

In this work we have deliberately grouped safety concerns into four distinct categories: (\textit{i})~\textit{open-world context}, (\textit{ii})~\textit{data and data set preparation}, (\textit{iii})~\textit{DNN characteristics}, and (\textit{iv})~\textit{analysis and evaluation}.
This specific categorization is based on the main sources of the corresponding concerns.
We can see a visualization of these categories in Figure~\ref{fig:structure}.
While \textit{open-world context} and~\textit{analysis and evaluation} are concerns for any AD component, \textit{data and data set preparation} and \textit{DNN characteristics} are both deep learning-specific categories.
As previous work has discussed~\cite{wang2023designing}, an important aspect of responsible AI development is sensitizing and communicating across different roles in organizations.
Just from the technical side we see the involvement of system engineers, machine learning engineers, data engineers, V\&V engineers, and safety engineers.
In addition, the documentation of concerns and corresponding stakeholders allows us to clearly outline interfaces between engineers in the development team.

\section{Open-World Context Concerns}
\label{sec:open}
Autonomous driving systems are deployed in an open world, which is a complex environment that evolves over time.  
This poses serious challenges on representing the ODD distribution with data independent of the algorithm that processes the data.
This section focuses on such safety concerns relating to the open-world context.

\subsection {Highly Limited Specifiability of Operational Design Domain}
\label{subsec:specifiability}

The ODD can be highly complex and therefore, it is not possible to specify it in arbitrary degrees of detail.
This is not limited to, but especially relevant for open-world contexts.
Let us consider one single road traffic scenario, \eg an intersection.
An abstract description (representation) includes possible elements of the scene, such as traffic lights and signs, lanes, road geometry, traffic participants and combinations thereof.
For a more detailed description, we focus on a video-based perception component working with RGB images.
Here, a detailed description of a scenario would additionally include information such as illumination, weather, appearance of above elements including properties such as color and texture, etc.
In both cases, the included information depends on the downstream use of the description: in the abstract description, the focus could be planning, where visual appearance is not of interest, while the second is focused on perception capabilities.

At some point, \eg when the non-semantic level of pixel distribution is reached, the combinatorial possibilities are practically infinite and detailed specification becomes impossible requiring a different approach, \ie specification on a different level as described above.
Hence, the ODD distribution $\mathcal{O}_{\mathcal{W}}$ can only be a coarse-grained and approximate high-level semantic description of the relevant part of the world. It is practically impossible to determine the set of all relevant properties $\mathcal{W}$.
Data is directly affected by this limitation, \ie its content cannot be specified in detail.
Instead, the ODD provides the means for determining the strategy for data acquisition, which in turn should allow for collection of \textit{unspecified} random lower-level details.
This could be achieved by recording data in different situations under various conditions in the real world as opposed to on the test track or synthetic data generation.

\subsection{Insufficient Coverage of the Open-World ODD}
\label{subsec:coverage}

AD is an open-world problem with practically infinite amount of variability.
In practice only a finite amount of data can be sampled that should cover the actual ODD distribution as well as possible.
This is difficult since the ODD distribution is not balanced, \ie some elements, conditions, or events will occur more often than others.
At the same time, rare events with high severity can have an equally large safety impact as frequent events with lower severity.
As an example for object detection, some object classes, or variants of the same object class, will appear so rarely in a data set, that the detector may not be able to predict them with the same accuracy as classes that appear (more) often. 
Naturally, data collected from the relevant domain, or generated synthetically in order to cover it, may (\textit{i})~not approximate the actual ODD distribution and (\textit{ii})~miss parts of the distribution.
As a result, there will always be a gap between the sampled data and the ODD.
An important part of safety engineering is to analyze and mitigate this gap.
A particular challenge for closing this gap is considering the tails of the distribution, especially heavy tails in which the probability of events are not exponentially bounded and thus cannot be neglected.

Note that coverage and specifiability may work hand in hand: resulting specification shows us what is known and allow us to subsequently derive corresponding coverage goals, \eg leveraging a systematization of visual corner cases~\cite{breitenstein2021corner}.	
For data coverage, we additionally need to consider -- as for the overall AD system  -- the unknowns and provide an argument why their contribution to the residual risk is sufficiently small. 
Approaches such as monitoring mechanisms help to uncover unknowns in the field and add them to a specification.
These approaches may require specific measures on system level as well as on organizational/management level.

\subsection{Distributional Shift Over Time}
\label{subsec:shift}

Deep learning relies on the fact that a DNN is trained and makes predictions on a stationary distribution.
However, there is a natural shift in the input space with which the model is confronted during its operational lifetime, \cf~Sec.~\ref{subsec:dist}. This is referred to as distributional shift over time, \eg~see~\cite{yao2022wild}.

This shift could increase the mismatch between the training sample distribution the model has been developed with, and the target sample distribution the model is confronted with during its operation. This, in turn, increases the risk of degraded model performance~\cite{vela2022temporal}.

The distributional shift can stem from different sources and can occur on different time scales.
For example, the weather may change depending on the season (summer vs. winter), the sensors might degrade over time due to aging, or other changes might occur due to cultural/technological evolution (such as fashion and new technologies).

\section{Data and Data Set Preparation}
\label{sec:data}

Deep learning is fueled by data which comes in many forms: (\textit{i})~input data that is used by a DNN during training and prediction, (\textit{ii})~labels for training and evaluation, (\textit{iii})~meta-labels that may be used for data set stratification and splits and (\textit{iv})~the construction of various datasets, \eg for training or various forms of testing~\cite{abrecht2021testing}.
In the following, we discuss concerns \wrt data, such as leakage and label quality.
While some of the data considerations have already been described in the context of the open world in~Sec.~\ref{sec:open}, here we focus on the particularities of data usage during development.
Our focus will be mostly on training aspects, as evaluation is discussed separately in~Sec.~\ref{sec:eval}.

\subsection{Leakage in the Development Cycle}
\label{subsec:leakage}

Deep learning relies on the assumptions that there is a characteristic ODD distribution, \cf~Sec.~\ref{subsec:coverage}, and that this distribution is stationary, \cf~Sec.~\ref{subsec:shift}.
For performance assessment, a standard assumption in practice is that there are separate datasets for training and evaluation, which are independent and identically distributed (i.i.d.) samples from the sensor data distribution.
Based on these datasets the generalization error is empirically determined.
In general, (\textit{i})~some assumption on the underlying distribution is needed and (\textit{ii})~if there is no further knowledge of requirements on the application, a statistical approach based on i.i.d. sampling is typically used, \cf~Sec.~\ref{subsec:risk}.
If the data sets suffer from leakage, the independence assumption between data sets cannot be met and therefore, there is a risk that performance evaluation is biased and unreliable.
Such leakage can come in various forms, such as 
(i) for a sequence, \eg (almost) identical images of the same sequence can be found in separate data sets,
(ii) while data sets are not necessarily the same, the data is recorded such that the independence assumption may be violated, \eg data from particular areas in a city.
As an example, in the Cityscapes dataset, each city is uniquely assigned to a single split~\cite{cordts2016cityscapes}.

To avoid this, a dedicated \textit{hold-out set} for evaluation can be used, which is not provided to developers to ensure that it remains independent and no information leaks into the development process. 
In practice, \eg in machine learning competitions~\cite{deng2009imagenet}, a test set is used that is withheld, but information is leaked though consecutive evaluation of models on the test set.
Recent work~\cite{recht2019imagenet} has shown that while this leakage may not necessarily lead to wrong model selection, performance evaluation can nevertheless be unreliable.

\subsection{Insufficient Labeling Quality}
\label{subsec:labeling}

In supervised learning, labeled data is the basis for training. 
Therefore, the quality of labels directly influences the performance and generalization capabilities of a DNN.
If labels are of low quality or wrong, systematic errors may be introduced in the model. 
In general, the reliability of a trained DNN decreases with a reduction in label quality.\footnote{Small amounts of label noise in training may be helpful for generalization, especially if uncertainties in ground truth are not explicitly reflected in labels.}
Note that label quality needs to be considered for all label sources, whether human annotation-based, sensor-based, algorithmically generated, or implicitly generated pseudo-labels.  

Since labels are usually an integral part of evaluation, label quality also directly increases the risk of unreliable evaluation.
Particularly, this might result in an over- as well as underestimation of the DNN's capabilities.
This may lead to issues in model selection and, even worse, in unexpected and insufficient performance in the field.

Label issues are a known problem in machine learning~\cite{DBLP:conf/nips/NorthcuttAM21} and also autonomous driving datasets. 
As an example, Kang \etal~\cite{kang2022finding} devise a tool to identify labeling errors and identify various types of errors in private and public datasets.
Reaching perfect labeling is an elusive goal: labels depend on context of the ODD and the labeling process is subject to specifications which are defined by humans and therefore subject to alignment considerations, \cf~Sec.~\ref{sub:misaligned}.

\subsection{Susceptibility to (Data) Biases}
\label{sub:bias}

As explained in Sec.~\ref{subsec:coverage}, the distribution of an open-world context is usually imbalanced.
What the distribution refers to depends on the application and the degree of detail considered, \cf~Sec.~\ref{subsec:dist}. It could be the ODD distribution $\mathcal{O}_{\mathcal{W}}$, \ie on a semantic level, or the sensor data distribution, \eg on raw pixels, $\mathcal{S}^{video}_{(\mathcal{X}, \mathcal{Y}|\mathcal{W})}$.
As an example of imbalance, there are usually fewer wheelchair users than walking pedestrians. 
Such imbalances will naturally reflect themselves in the sampled data and can be seen as data biases.
Additional bias may be introduced, intentionally or unintentionally, via concepts, processes, and activities determined and performed by people involved in the development. 
For example, the geographic area selected for the operation of the system has an effect on data distribution, or supplementing data for underrepresented cases may affect the performance on other cases.
In this subsection, we focus on data and dataset preparation \wrt desired properties of the model or the system in which it is applied and \textit{not} \wrt the ODD distribution.

Properties that a DNN exhibits are not specified but emerge after training.
The bias in data affects these properties. While the developers intend that introduced biases have a positive effect, \eg that detection also works well on underrepresented object classes, they could also have negative effects.
For example, an analysis on synthetic data shows that different attributes of the generated pedestrian heavily impact the detection performance of a pedestrian detector~\cite{grau2022variational}.
As DNN properties may be misaligned to developer intentions, \cf~Sec.~\ref{sub:misaligned}, it may be unclear how to detect the manifestation of the data biases on properties, \eg unwanted side-effects.
Therefore, they can remain unidentified by tests and analyses.
This needs to be considered in the evaluation of system's residual risk.

Even though the underlying safety problem, discussed above, is the generalization to unseen data, the socially relevant aspect of fairness needs to be emphasized here. 
It is an important requirement for systems with social interactions, that they do not systematically discriminate against groups of people.
This might be particularly challenging if groups are underrepresented in the ODD distribution.
However, it is always possible to consider a slice-based evaluation, \cf~Sec.~\ref{sub:safetyaware}, for a detailed analysis and evaluation for these groups.

\subsection{Domain Mismatch}
\label{sub:mismatch}

We discussed above that changes to the distribution are a safety concern and that distributional shift over time is a property of the open-world context, \cf~Sec.~ \ref{subsec:shift}.
However, distributional shifts may also occur due to the way the data is captured, processed, and provided to the system.
As an example, consider a camera image: the sensor data distribution differs for data recorded with different cameras, different normalizations, post-processing using some form of augmentation, or for synthetically generated data.
This difference of data sources may result in a distributional shift between the development data and the data that the DNN processes during system operation.
We call such sources of distributional shift \textit{domain mismatch}.

Domain mismatch can occur in various forms and might stem from many different sources. 
It can occur on a semantic level, \ie $\mathcal{O}_{\mathcal{W}}$, due to missing or underrepresented objects or environmental conditions, \eg in different geographic regions.
It can also occur on a sensor data level, \ie $\mathcal{S}_{(\mathcal{X}, \mathcal{Y}|\mathcal{W})}$, such as pixels and point clouds, due to usage of different sensor types between data sets or due to synthetic data or data augmentation that is not representative of the target domain.
If there is a domain mismatch between the training data and the target domain, the model might not be able to learn the appropriate features and concepts present in the target domain and unwanted biases might be introduced to the model, \cf Sec.~\ref{sub:bias}.
As an example, experiments on OOD detection in~\cite{hell2021monitoring} also clearly show that domain mismatches – in this case due to weather conditions – can heavily impact the performance, here of a depth estimation network.
If there is a domain mismatch between the test data and the target domain, the model performance in the field cannot be accurately estimated.
This wrong estimation remains uncovered if the network behavior under target population is not explicitly analyzed, \cf~Sec.~\ref{sub:misaligned}.
However, it needs to be noted that the usage of data collected by (potentially many) different sensor types or mounting positions can be useful if utilized carefully.
For example, for training data this can be seen as a natural form of data augmentation and for test data this can be seen as a form of robustness testing.

\section{DNN Characteristics}
\label{sec:DNN}

DNNs are universal function approximators, \ie they can fit any possible function. The high-dimensionality of their input and output space results in a high-dimensional space of possible functions, \cf~Sec.~\ref{sec:background}.
The specific function is determined via the training process, \ie fitting the model parameters, and depends on various factors such as the architecture, losses, and the training (and validation) data.
Model parameters determine the features the DNN extracts from an input and how these are leveraged to determine a prediction.
Hence, extracted features and also resulting properties of the DNN are mostly not predetermined by designers, but rather learned in the training process.
Most DNN features as well as properties can neither be formally analyzed nor interpreted by humans.
In this subsection, we discuss the safety concerns arising from these characteristics.

\subsection{Idiosyncrasies of the Model Building Process}
\label{subsub:trainidio}

As a prerequisite for safety, a DNN needs to provide sufficient functional performance across the ODD distribution.
Regretfully this may be sometimes at odds with safety concerns.
A simple example considering adversarial robustness is that an adversarial training may negatively impact performance on nominal data or reduce robustness to some corruptions~\cite{yin2019fourier}.
Another example is the consideration of rare, yet critical cases versus the bulk of the distribution.

In order to achieve a given level of functional performance some design decisions may be taken. 
While general training issues such as overfitting, underfitting or hyper parameters selection are well-documented in machine learning literature, \eg~\cite{hastie2009elements}, we want to highlight additional particular issues from a safety perspective. 
These include (\textit{i})~training data selection, \eg overemphasis of rare yet important cases or data slices, pretraining or introducing synthetic data, (\textit{ii})~the formulation of a (safety-focused) loss function including weighted composition of several losses and  (\textit{iii})~the choice of model architecture.
As an example of (\textit{ii}), a standard loss used in classification tasks is cross-entropy even though it is agnostic to the fact that certain misclassifications are more safety-critical than others.
In order to address (\textit{iii}), designers may expose additional outputs from perception models for downstream use in fusion or planning~\cite{pmlr-v205-shao23a}.
Note that while one can argue that a loss is a partial specification of intended behavior and thus closely related to Sec.~\ref{subsec:specifiability}, training a model (and its convergence) may require to adapt the loss function.
Concretely, it is more safety-critical to classify a person as road, than classifying a bus as a truck.

Design decisions require justification since idiosyncrasies introduced in the model building process can have detrimental effects on the behavior of the network, not all of which might be unveiled by analysis and evaluation. 
It needs to be argued why decisions were sensible or necessary by providing qualitative and quantitative evidence regarding possible negative effects, \eg from dedicated analysis and evaluation.

\subsection{Uncertainty Quantification Challenges}
\label{sub:UQ}

Reliable uncertainty quantification regarding the predictions of a DNN is key in safety-relevant applications as it enables informed decision-making on a system level, such as degrading functionalities.
However, acquiring accurate uncertainty quantification can be challenging.
There are different sources of uncertainty that convey different information and require different quantification methods~\cite{huellermeier2021aleatoric, gawlikowski2023survey}.
For example, a single DNN model can report its aleatoric, \ie data, uncertainty, when trained with a corresponding loss~\cite{kendall2017uncertainties}.
In contrast, determining epistemic, \ie model, uncertainty requires additional sources of knowledge, \eg via dropout variational inference~\cite{kendall2017uncertainties}.
These uncertainties can be included in the design of the model and used either during field operation, \ie runtime or online measures, or in the development cycle, \ie offline measures.
Whether online or offline measures are suitable depends on factors such as resource demands of the quantification method or the necessity of involving humans for further analysis.
However, some uncertainties may not be quantifiable at the level of the model or even the system itself, \eg ontological uncertainty~\cite{gansch2020design}.
Some safety concerns \wrt uncertainty quantification originate from the domain and the context.
However, most concerns with uncertainty quantification result from using DNNs as further discussed in the following.

Apart from the different types of uncertainties, it is important to realize that when uncertainty is quantified via DNNs, the resulting estimations will be subject to errors, \ie the same generalization issues as for other predictions of DNNs.
It is also well known that confidence estimations of DNNs are typically poorly calibrated and require an explicit calibration~\cite{guo2017calibration}. Moreover, calibration will degrade for out-of-training-distribution data ~\cite{hein2019relu}. 
Also noteworthy is the so-called softmax confidence commonly used for classification and by many object detection algorithms such as non-maximum suppression.
These softmax values cannot be interpreted as probabilities of a model's prediction being correct, as they are based on a ``closed world'' assumption of fixed set of classes, which is in contrast to the open-world context.

\subsection{Brittleness}
\label{sub:brittle}

DNNs exhibit brittleness meaning that changes to the input -- that do not change the local semantics -- may cause large changes in the prediction~\cite{zheng2016improving}. As an example, overlaying an object onto an existing image can change the detected class showing brittleness to contextual cues~\cite{wang2017visual}: a monkey with an overlayed guitar is suddenly detected as a person.
All kinds of natural input changes including illumination, weather conditions and sensor noise, or targeted attacks such as adversarial examples~\cite{hendrik2017universal} may cause such an effect.

Brittleness may also occur across consecutive predictions in a data stream, \eg consecutive frames in a video~\cite{zheng2016improving}. As such, spatio-temporal instability is a manifestation of brittleness. 
As an example, let us consider a typical object detection task in the context of an AD system. Even though there are only small changes in the input over a short video sequence, an object may be detected sporadically, or its associated confidence may vary significantly. Such brittleness poses a challenge for receiving components, like a tracker or a fusion component.

\subsection{Incomprehensibility}
\label{subsub:incomprehensible}

A DNN's strength to solve highly complex tasks comes with the incomprehensibility of how it derives a prediction~\cite{houben2022inspect}. This largely stems from two sources:
Firstly, well-understood hand-crafted feature extractors are replaced by self-learned ones that are tuned during the training process. Those are mostly counterintuitive to humans - especially if the DNN at hand operates on a high-dimensional and non-semantic input space (\eg pixels of an image).
Secondly, large amounts of neurons in the DNN as well as non-linearities introduced through activation functions additionally impede the understanding of the connection between extracted features and output.
From a safety perspective, we aim to understand sources of errors and argue their mitigation in a safety argumentation based on evidence. 
The incomprehensibility of a model and its functional insufficiencies is a safety concern as it limits this safety argumentation. 
It also reduces the evaluation capabilities to mostly statistical tests of a black box.

\section{Analysis and Evaluation}
\label{sec:eval}

Analysis and evaluation of a DNN's appropriateness is an indispensable part of using deep learning responsibly within a system and the corresponding environment~\cite{hutchinson2022evaluation}.
While in best-effort systems the focus may be myopically on optimizing a DNN with a few metrics on a standard benchmark~\cite{rostamzadeh2021thinking}, for DNNs that are part of a perception component in a safety-critical system  various objectives, factors, and issues need to be considered.
As such it is evident why safety standards require that the function to be applied in a safety-critical setting is understood concerning its functional and output insufficiencies~\cite{sotif}. 
The safety of a component has to be argued in a safety argumentation (of the enclosing system) and this requires strong evidence in the form of thorough evaluations.
Note that this requires an application-specific evaluation of the DNN in addition to its learner-specific evaluation~\cite{hutchinson2022evaluation}.
Additionally, we need to consider evaluation gaps~\cite{hutchinson2022evaluation}, \eg that evaluations and corresponding measurements actually target the concept (safety concern) to be addressed, so-called \textit{concept validity}~\cite{9780058}.
While this category applies to any perception component, we particularly focus on analysis and evaluation of DNNs in the following.

\subsection{Misalignment of Intention, Specification, and Implementation}
\label{sub:misaligned}

In the development of open-context systems, Stellet~\etal\cite{stellet2019formalisation} characterized various distinctive deviations that can emerge between the (\textit{i}) required (intended), (\textit{ii}) specified, and (\textit{iii}) implemented behavior of a system. 
The 3-circles model shows us that there may be concerns with respect to the three behaviors, and we already discussed this \wrt (\textit{ii}) specifications Sec.~\ref{subsec:specifiability} and (\textit{iii}) the implementation \ref{subsub:incomprehensible}.
However, as shown in~\cite{stellet2019formalisation} deviations also occur in the relation between the behaviors.

First, an explicit specification of the intended properties of a DNN is elusive due to the complex and dynamic nature of the problems for which such algorithms are usually used, and the environments they are deployed in, \cf~Sec.~\ref{sec:open}. Those properties are rather implicitly given by data and training aspects, \eg DNN architecture. 
In fact, not requiring an explicit specification is actually a virtue making these algorithms so suitable for problems that cannot be specified in detail. But, as mentioned in~Sec.~\ref{subsec:specifiability}, the ODD distribution, and therefore the development data, is only partially specifiable. This leads to misalignments between intentions and specification.

Second, unlike rule-based algorithms which are explicitly implemented to perform a specific task, the functionality of a DNN is implemented implicitly. 
DNNs are mainly black boxes defined by training, \cf~Sec.~\ref{sec:DNN}.
Some specification of training data, losses, and architecture are possible, yet do not provide a full specification of the resulting model.
This results in an inevitable and well-known misalignment between intentions and the implementation, which has previously been described as the ``underspecification'' problem~\cite{d2020underspecification}.
One of the possible consequences is that two different implementations can show the same performance based on their respective loss functions, yet have completely different, non-obvious functional and output insufficiencies.

Third, testing a trained DNN suffers from the same issues mentioned above, since analysis relies on data and DNN properties cannot be analytically derived. 
Therefore, analysis and evaluation will also be subject to incompleteness. This means that not all undesirable or missing desirable model properties can be uncovered.
Hence, this needs to be considered when evaluating the residual risk of the system.

\subsection{Safety-Agnostic Performance Evaluation}
\label{sub:safetyaware}

In general, an evaluation shall provide trustworthy and transparent estimates of field performance considering the ODD distribution $\mathcal{O}_{\mathcal{W}}$,~\cf~Sec.~\ref{subsec:dist}.
DNNs are usually evaluated using average metrics such as false positive and false negative rates, mean squared error, mean intersection over union, etc.
This  focus on a model's average performance on a (test) data set is not sufficient for safety-relevant applications.
Functional and output insufficiencies may remain uncovered if the performance is only considered in this restricted sense~\cite{rostamzadeh2021thinking}. 
This is especially true if test sets heavily contain data samples from the body of the distribution.  
Moreover, in standard evaluation, predictions are compared to the ground truth irrespective of their particular relevance for a task, in our case the driving task~\cite{philion2020learning}. 
As an example, for a self-driving car nearby objects are usually more safety relevant than faraway objects~\cite{Lyssenko2022}. 
If all objects are equally weighted in an average evaluation metric, the performance \wrt safety may be underestimated. 

A strong safety argumentation can only be achieved by performing a thorough evaluation.
This necessitates the creation of safety-aware performance metrics, \eg~\cite{Lyssenko2022,varghese2021unsupervised}, that are better aligned with the intended and required behavior of a DNN, \ie properties expected implicitly or explicitly, \cf Sec.~\ref{sub:misaligned}, and may be based on domain-, application-,  and system-specific knowledge as described below.
This may also require the construction of various train and test sets and slices thereof, that do not need to be i.i.d. \wrt the domain~\cite{hutchinson2022evaluation, rostamzadeh2021thinking}, and corresponding safety-aware performance metrics~\cite{breck2017ml, chen2019slice, abrecht2021testing}.
An example is the construction of robustness test sets to analyze how the network performs and reports its uncertainty in novel situations, \cf~Sec.~\ref{sub:UQ}.
As another example, Lyssenko~\etal\cite{Lyssenko2022} describe the construction of a test set based on relevant pedestrian interactions and evaluating temporal instability over consecutive predictions.
As a final example, inference latency on the target platform is a further relevant property of DNNs as late predictions can result in the same insufficiencies as false predictions.
While previous work has even proposed to group such resource considerations as an individual source of (efficiency) insufficiencies~\cite{saemann2020strategy}, we see these as further safety-relevant properties that need to be evaluated properly.

We also need to make sure that test sets measuring performance are not too easy and that difficult tail cases are not hidden by an average evaluation with a large number of easy cases~\cite{rostamzadeh2021thinking, metzen2023identification}.
As an example, when an object detection test set is constructed with an overemphasis on large and unoccluded objects, the estimation of the mean performance will be an unreliable measure of the performance in field.

\subsection{Unknown Predictions for Rare Cases in the Long Tail}
\label{subsec:rare_cases}

The ODD distribution relevant for the DNN is long- and potentially even heavy-tailed, \cf~Sec.~\ref{subsec:coverage}.
Therefore, it is desirable to have a dedicated evaluation of the quality of DNN predictions in rare cases in the tail. 
The problem, however, is that due to the rareness of such cases, these may be missing from or underrepresented in the data.\footnote{Please note that this includes intra-class instances, \eg a strange and rare form of car.}
The complexity and openness of the ODD makes it impossible to anticipate all rare cases -- especially at a non-semantic level, \cf~Sec.~\ref{subsec:coverage}. 
However, their contribution to the residual risk needs to be determined according to the ISO~21448.

Remember that the term \textit{case} has different meanings depending on the level of abstraction; for example, a case could be on a higher level of abstraction, such as a driving situation, or a on a low level, such as the change of a patch in an image (\ie a combination of pixels). Additionally, not every rare case is difficult, even if not seen during development. Moreover, some cases are inherently hard and not necessarily learnable by a single sensor modality~\cite{jiang2022improving}, such as the detection of a strongly occluded pedestrian.

\section{Discussion and Related Work}
\label{sec:related_work}

In a previous work, some of the authors introduced safety concerns~\cite{willers2020safety}.
This included the description of the problem, relation to safety engineering, and the separation of problem and solution space.
In this work, we refine the previous nine safety concerns and provide a mapping between previous and new safety concerns and categories in Table~\ref{tab:old_current_concerns_mapping}. 

\begin{table}[h!]
	\renewcommand{\arraystretch}{1.3}
	\caption{Map between presented safety concerns and previous work~\cite{willers2020safety}.}
	\label{tab:old_current_concerns_mapping}
	\centering
	\begin{tabular}{p{35mm}p{45mm}}
		\hline	
		\textbf{Previous Safety Concern~\cite{willers2020safety}} & \textbf{Revised Safety Concern (this paper)} \\
		\hline
		\multicolumn{2}{c}{\textbf{Same}}\\
		\hline
		\hline
		Distributional shift over time~(\mbox{SC-2}) & Distributional shift over time~(Sec.~\ref{subsec:shift}) \\
		\hline	
		Incomprehensible behavior~(\mbox{SC-3}) & Incomprehensibility~(Sec.~\ref{subsub:incomprehensible}) \\
		\hline	
		Unknown behavior in rare critical situations~(\mbox{SC-4}) & Unknown predictions for rare cases in the long-tail~(Sec.~\ref{subsec:rare_cases}) \\
		\hline	
		Brittleness of DNNs~(\mbox{SC-6}) & Brittleness~(Sec.~\ref{sub:brittle}) \\
		\hline	
		Inadequate separation of test and training data~(\mbox{SC-7}) & Leakage in the development cycle~(Sec.~\ref{subsec:leakage}) \\
		\hline	
		Dependence on labeling quality~(\mbox{SC-8}) & Insufficient labeling quality~(Sec.~\ref{subsec:labeling}) \\
		\hline	
		\multicolumn{2}{c}{\textbf{Extended}}\\
		\hline
		\hline
		\multirow{4}*{\parbox{2cm}{Data distribution is not a good approximation of real world~(\mbox{SC-1})}}&
		Highly limited specifiability of operational design domain~(Sec.~\ref{subsec:specifiability})\\
		& Insufficient coverage of the open-world ODD~(Sec.~\ref{subsec:coverage}) \\
		& Susceptibility to (data) biases~(Sec.~\ref{sub:bias}) \\
		&  Domain mismatch~(Sec.~\ref{sub:mismatch}) \\
		\hline	
		Unreliable confidence information~(\mbox{SC-5}) &  Uncertainty quantification challenges~(Sec.~\ref{sub:UQ}) \\
		\hline	
		Insufficient consideration of safety in metrics~(\mbox{SC-9}) & Safety-agnostic performance evaluation~(Sec.~\ref{sub:safetyaware}) \\
		\hline
		\multicolumn{2}{c}{\textbf{New}}\\
		\hline
		\hline
		& Idiosyncrasies of the model building process~(Sec.~\ref{subsub:trainidio}) \\
		\hline	
		& Misalignment of intention, specification, and implementation~(Sec.~\ref{sub:misaligned})\\
		\hline
	\end{tabular}
\end{table}

As shown in the table, we group the concerns based on their update status, \ie whether a concern (\textit{i}) stayed the \textit{same}, (\textit{ii}) was \textit{extended}, or (\textit{ii}) was added, \ie is \textit {new}, based on feedback from involved engineers.
In particular, while in the original work by Willers~\etal~the training process that leads to DNN weights was seen as a black box, we explicitly include it since the knowledge of 
idiosyncrasies in the model building process may require additional analysis and evaluation, \eg whether certain assumptions used in data augmentation are actually valid or whether a particular selected loss function does not generate issues.

We further decided to include a particular concern for alignment of intention, specification, and implementation.
While one can argue that in particular intention and requirement specification are part of systems and safety engineering and should be outside the DNN scope, the peculiarities of deep learning and corresponding specifications that are not specifically implemented but implicitly created by training are particularly challenging.
Relation to specifiability
Therefore, alongside several recent works that have emphasized the importance of alignment between high level intentions and emergent properties of resulting machine learning models~\cite{kuwajima2020engineering,hendrycks2021unsolved}, we introduce alignment as a safety concern to highlight the necessity for interaction between systems, safety, and machine learning engineers.

Finally, safety concerns may be (partly) overlapping.
This is not an issue: If a mitigation can be identified, it can address all concerns in this overlap.
Let us consider distributional shift in Sec.~\ref{subsec:shift} and~\ref{sub:mismatch}. Some methods, \eg drift detection, may be shared across the concerns.
However, there is a difference between addressing shifts of sensor data and a changing context, \eg that the latter cannot be controlled because it is outside of the technical system.
In such cases, we rather add an additional concern such that safety engineers and mitigation developers are aware of this difference, rather than grouping both sources of distributional shift in a larger, less nuanced, joint concern.

Note that the original paper on safety concerns~\cite{willers2020safety} sparked further discussions and refinements: Within the German publicly funded project "Safe AI for Automated Driving", safety concerns were used to structure the safety argumentation, and, to develop mitigation methods.\footnote{\url{https://www.ki-absicherung-projekt.de/en/}}
Houben~\etal~\cite{houben2022inspect} survey practical methods
for AI Safety considering topics such as data, training, and verification and validation. Additionally, a diverse set of novel contributions in Safe AI for Automated Driving is presented. Hence, in contrast to the safety concerns discussed in this paper, the focus of the book is on mitigation methods that can be used to address safety concerns.
Mock~\etal describe a safety argumentation for DNNs by systematic consideration of safety concerns and corresponding mitigation~\cite{mock2021integrated}.
Condurache~\cite{condurachesafety} leverages the concepts of the original safety concerns and compares them to generalization considerations. Concretely, the author relates classical generalization bounds with corresponding parameters such as dataset size, to individual safety concerns. This links possible sources in design and training of DNNs to resulting safety concerns. In contrast, the updated safety concerns in this work and their corresponding categorization show that issues with DNN safety do not only originate in design and training, but also stem from the open-world context as well as challenges in analysis and evaluation.
S\"amann~\etal~\cite{saemann2020strategy} attribute five insufficiencies to DNNs, which they define as systematic and latent weaknesses. These are \textit{lack of generalization, robustness,  explainability,  plausibility}, and \textit{efficiency}. Additionally, mechanisms are  introduced for mitigation and metric categories for evaluating the effectiveness of mitigations. 
The insufficiencies discussed in the paper focus on missing properties of DNNs. In contrast, this work focuses on underlying sources of insufficiencies.
Furthermore, the present work distinguishes different categories, which identify that safety concerns are not only due to using DNNs, but also due to the open-world context, and analysis and evaluation challenges.

Schwalbe~\etal~\cite{schwalbe2020structuring} consider DNN insufficiencies to be intrinsic properties of such algorithms, which negatively impact the safety of the corresponding system. As such, they identify the following specific insufficiencies: \textit{black-box nature}, \textit{simple performance issues}, \textit{incorrect internal logic}, and \textit{instability}. Based on these, the authors break down safety requirements, which need to be fulfilled for a sufficient absence of risk, and identify two types of evidence necessary for arguing sufficient safety, namely, ``detection and measurement'' and ``prevention and mitigation''. The main focus of this work lays on safety argumentation and its structure. In contrast the work at hand focuses on comprehensive description of the problem space.

Kuwajima~\etal~\cite{kuwajima2020engineering} discuss engineering problems in machine learning systems. 
The paper focuses on a lack of requirements specification and design specification, \cf our concern on misalignment in Sec.~\ref{sub:misaligned}, lack of interpretability and robustness of DNNs, which we detail more fine-granularly in Sec.~\ref{sec:DNN}. 
For these concerns, the paper discusses related work in mitigations and identifies current gaps.
In contrast, we detail on various further concerns that need to be considered for engineering DNNs in the AD domain, where \eg challenges of the open-world context need to be considered (Sec.~\ref{sec:open}).

Previous work has investigated leveraging classical safety engineering methods for machine learning components~\cite{rismani2023plane}.
The interviews in the above-mentioned paper as well as the one by Martelaro~\etal~\cite{martelaro2022exploring} indicate that engineers see a need for safety engineering and identifying risk in applying machine learning.
Furthermore, in the latter work, the interviewed engineers see that (among other things) there is a need for better collaboration, better tooling, and a need for better understanding of capabilities, but also limitations of machine learning.
We can conclude from these works that no matter the safety approach to be used, safety engineers need a good understanding of safety concerns that may lead to hazardous behavior of the system as shown in~Fig.~\ref{fig:intro}, and therefore, introduce risk of harm.
This is the basic motivation for introducing safety concerns as they are a suitable structuring element aligned with safety standards.

Hendrycks~\etal~\cite{hendrycks2021unsolved} discuss unsolved problems for machine learning safety in general. They categorize these problems into 4 categories: \textit{robustness, monitoring, alignment}, and \textit{external safety}.
Some of the considerations are similar. We can see that robustness addresses concerns such as brittleness described in~Sec.~\ref{sub:brittle}. 
However, the problems discussed in the paper are both within problem space (\eg robustness) as well as in solutions space (\eg monitoring).
Safety concerns, however, consider specifically the problem space, while the solution space is addressed with separate mitigations. This separation is important because mitigations such as monitoring can address various concerns in problem space.
Additionally, the focus of the safety concerns introduced in the work at hand is deep learning for AD systems. So, topics such as alignment are particularly focused on the respective operational design domain rather than all possible alignment concerns.

Two recent works~\cite{rostamzadeh2021thinking, hutchinson2022evaluation} focus on evaluation of machine learning models. 
Rostamzadeh~\etal~\cite{rostamzadeh2021thinking} detail why simple i.i.d. test sets with standard metrics are not sufficient for evaluation.
Hutchinson~\etal~\cite{hutchinson2022evaluation} extend this work and identify six (often implicit) assumptions that simplify the model evaluation task, but may not be valid in the application domains, \eg that failure cases can all be treated the same.
The paper discusses eight corresponding evaluation gaps that -- if not addressed -- may render evaluation unreliable.
Since evaluation and corresponding analysis is such a vital part in machine learning practice, we introduce a complete concern category, \cf~Sec.~\ref{sec:eval}, to highlight its importance and focus on corresponding concerns for our application domain.
Wang~\etal~\cite{wang2023designing}~discuss challenges in addressing responsible AI concerns in industry and conduct a survey of practitioners with a corresponding analysis.
Again, the authors find that responsible evaluation is an important factor.
The paper discusses responsible prototyping as an option, as a form of online evaluation.
It also proposes the concept of a lens, \ie to focus on responsible AI.
Safety concerns are such a lens that allows developers to focus on safety and to communicate across different roles in organizations.

NIST has recently provided a comprehensive risk management framework for AI that addresses risks and the trustworthiness of AI systems~\cite{NIST2023}.
They do not only discuss safety, but also other risks such as security and resilience.
Since it is a general framework, it is also domain- and application-agnostic.
The core of the framework is decomposed into four functions: \textit{map}, \textit{measure}, \textit{manage}, and \textit{govern}.
Our work mainly concerns the map function, which is described as the ``context is recognized and risk related to the context are identified''~\cite{NIST2023}. 
Relating to the NIST framework, the main point of our work could be seen as the \textit{map} function discussed above, \ie analyzing the usage of DNNs in AD systems and identifying corresponding domain- and application-specific safety concerns that contribute to system-level risks.

Also important to mention is that there are several standards in the context of DNNs and AD systems. 
Most notably, we already referred to the safety of the intended functionality (SOTIF)~\cite{sotif} and showed how our safety concerns support the concept, as shown in Fig.~\ref{fig:intro} and described in Sec.~\ref{subsec:sc_intro}.
This paper tries to further inform DNN safety practitioners, such that safety concerns can be considered in upcoming standards, \eg on safety and artificial intelligence in road vehicles~\cite{iso8800}.

Even though we focus on the AD domain and safety, we also mentioned works from other domains, \eg~\cite{hutchinson2022evaluation, hendrycks2021unsolved, breck2017ml, d2020underspecification}.
Similarly, safety mitigation may also be inspired by defenses from the security domain~\cite{deng2021deep}.
This is because also in non-safety critical domains, there may be undesirable consequences from using deep learning-based systems.
As such, there is potential to learn across domains to identify and mitigate insufficiencies of DNNs.

Finally, our focus has been on perception in autonomous driving and thus models trained in (self-) supervised fashion.
For AD systems considering end-to-end approaches, additional learning paradigms and challenges need to be considered~\cite{atakishiyev2024safety}.

\section{Conclusion}\label{sec13}

This paper discussed a systematic and comprehensive approach, so-called safety concerns, that can be leveraged as a suitable structuring element for safety engineers.
Safety concerns are the result of a domain-specific analysis of the problem space that arises when deep learning is leveraged in safety-related tasks, such as perception in AD systems.
The concerns identified in this paper are for the context of perception tasks in AD systems.
We refined and extended safety concerns from previous work and introduced a categorization to better understand, structure, and communicate the problem space, \eg to help cross-functional teams in addressing safety concerns.
We identified and detailed on fourteen safety concerns across four categories relating to (\textit{i})~the open-world context the automated vehicle operates in, (\textit{ii})~data and data set preparation, (\textit{iii})~DNN characteristics, and (\textit{iv})~the analysis and evaluation of the DNNs within their operational design domain.
Our main motivation is to structure the problem space (``What are the concerns?'') to guide future work on the solution space, \ie mitigations (``How can concerns be addressed?'').
This allows developers of mitigations to not only focus on how a mitigation works, but clearly outline what underlying safety concerns can be addressed and to which extent.

\bibliographystyle{IEEEtran}
\bibliography{refs}

\end{document}